\documentclass[acmsmall, nonacm]{acmart}

\AtBeginDocument{%
  \providecommand\BibTeX{{%
    \normalfont B\kern-0.5em{\scshape i\kern-0.25em b}\kern-0.8em\TeX}}}

\setcopyright{cc}
\acmJournal{TACCESS}
\acmYear{2023} 
\acmVolume{16} 
\acmNumber{4} 
\acmArticle{25} 
\acmMonth{12}
\acmDOI{10.1145/3636513}

\usepackage{acronym}

\begin{document}

\title{Measuring the Accuracy of Automatic Speech Recognition Solutions}

\author{Korbinian Kuhn}
\email{kuhnko@hdm-stuttgart.de}
\orcid{0009-0005-1296-4987}
\affiliation{%
  \institution{Stuttgart Media University}
  \streetaddress{Nobelstraße 10}
  \city{Stuttgart}
  \country{Germany}
  \postcode{70569}
}

\author{Verena Kersken}
\email{kersken@hdm-stuttgart.de}
\orcid{0009-0000-5007-6327}
\affiliation{%
  \institution{Stuttgart Media University}
  \streetaddress{Nobelstraße 10}
  \city{Stuttgart}
  \country{Germany}
  \postcode{70569}
}

\author{Benedikt Reuter}
\email{reuter@hdm-stuttgart.de}
\orcid{0009-0004-4618-376X}
\affiliation{%
  \institution{Stuttgart Media University}
  \streetaddress{Nobelstraße 10}
  \city{Stuttgart}
  \country{Germany}
  \postcode{70569}
}

\author{Niklas Egger}
\email{egger@hdm-stuttgart.de}
\orcid{0009-0003-9866-2175}
\affiliation{%
  \institution{Stuttgart Media University}
  \streetaddress{Nobelstraße 10}
  \city{Stuttgart}
  \country{Germany}
  \postcode{70569}
}

\author{Gottfried Zimmermann}
\email{zimmermanng@hdm-stuttgart.de}
\orcid{0000-0002-3129-1897}
\affiliation{%
  \institution{Stuttgart Media University}
  \streetaddress{Nobelstraße 10}
  \city{Stuttgart}
  \country{Germany}
  \postcode{70569}
}
\renewcommand{\shortauthors}{Kuhn and Kersken, et al.}

\begin{abstract}
For d/Deaf and hard of hearing (DHH) people, captioning is an essential accessibility tool. Significant developments in artificial intelligence (AI) mean that Automatic Speech Recognition (ASR) is now a part of many popular applications. This makes creating captions easy and broadly available - but transcription needs high levels of accuracy to be accessible. Scientific publications and industry report very low error rates, claiming AI has reached human parity or even outperforms manual transcription. At the same time the DHH community reports serious issues with the accuracy and reliability of ASR. There seems to be a mismatch between technical innovations and the real-life experience for people who depend on transcription. Independent and comprehensive data is needed to capture the state of ASR. We measured the performance of eleven common ASR services with recordings of Higher Education lectures. We evaluated the influence of technical conditions like streaming, the use of vocabularies, and differences between languages. Our results show that accuracy ranges widely between vendors and for the individual audio samples. We also measured a significant lower quality for streaming ASR, which is used for live events. Our study shows that despite the recent improvements of ASR, common services lack reliability in accuracy.
\end{abstract}





\newacro{DHH}[DHH]{d/Deaf and hard of hearing}
\newacro{ASR}[ASR]{Automatic Speech Recognition}
\newacro{WER}[WER]{Word Error Rate}
\newacro{WWER}[WWER]{Weighted Word Error Rate}
\newacro{CER}[CER]{Character Error Rate}
\newacro{MER}[MER]{Match Error Rate}
\newacro{WIL}[WIL]{Word Information Lost}
\newacro{ACE}[ACE]{Automated-Caption Evaluation}
\newacro{WPM}[wpm]{words per minute}
\newacro{WCAG}[WCAG]{Web Content Accessibility Guidelines}
\newacro{WAI}[WAI]{Web Accessibility Initiative}
\newacro{FCC}[FCC]{Federal Communications Commission}
\newacro{AI}[AI]{artificial intelligence}
\newacro{NAD}[NAD]{National Association of the Deaf}
\newacro{E2E}[E2E]{end-to-end}
\newacro{RTF}[RTF]{real-time factor}
\newacro{UPWR}[UPWR]{unstable partial word ratio}
\newacro{ESB}[ESB]{End-to-End Speech Benchmark}
\newacro{API}[API]{application programming interface}
\newacro{ESL}[ESL]{English as a second language}
\newacro{SDK}[SDK]{software development kit}
\newacro{GenAm}[GenAm]{General American English}
\newacro{RP}[RP]{Received Pronunciation}
\newacro{CART}[CART]{Communication Access Realtime Translation}

\maketitle

\section{Introduction}

Transcription is important to make spoken language accessible to \ac{DHH} individuals. With the rapid development of \ac{AI}, astounding results in the field of \ac{ASR} are reported, particularly in terms of accuracy \cite{Whisper2022}. Fully automated captioning is now included in online meeting tools, video streaming platforms, and presentation software. \ac{ASR} is scalable, cheap and offers a solution for the increasing amount of digital content and resulting demand for accessibility. But transcripts and captions generated using \ac{ASR} are only useful for those who depend on them, if they have very high levels of accuracy. This requirement is seemingly fulfilled - Scientific publications report extremely low error rates on transcription tasks \cite{Baevski2020}, and vendors produce “state-of-the-art accuracy” \cite{Google2023}. At the same time, the \ac{NAD} has filed a petition in the US to improve \ac{ASR} based captioning because people have reported serious issues with the correctness, timing and completeness of captions \cite{NADPetition2019}. There seems to be a mismatch between reports of \ac{AI} research and the lived experience of the \ac{DHH} community \cite{Kawas2016, Kafle2016, Butler2019, NDC2020}. The \ac{NAD} proposes to regulate the use of \ac{ASR} and the applied metrics to specify accuracy of captions and transcripts. We therefore need independent, descriptive data about the accuracy of \ac{ASR}, and whether the achieved accuracy level is sufficient to make captions accessible. 

The accuracy of \ac{ASR} is commonly measured using the \ac{WER}, which represents the amount of errors in a transcript compared to a sample solution. One main aim of research and development in \ac{ASR} is to reduce the \ac{WER} of datasets through the application of new technological innovations, e.g. \ac{E2E} models or self-supervised learning. Some of these tests produce very low error rates that even outperform those achieved through manual transcription \cite{Zhang2022}. But these high reported accuracy rates might not be a reliable indicator of the general quality of \ac{ASR}. Training and optimisation for specific datasets bear the risk of overfitting, and the obtained results might not be a good indicator of performance on novel datasets \cite{Geirhos2020}. Many sources show \ac{ASR}’s volatility with different datasets: \ac{ASR} performs differently for male or female speakers, shows racial biases or varying accuracy rates for speakers with different accents \cite{Tatman2017, Koenecke2020, Cumbal2021, Tadimeti2022, Speechmatics2023}. The studies aim to show that \ac{ASR} has specific biases that often affect speakers of underrepresented groups, but are not concerned with the overall accuracy of the engines. This baseline data is still missing. 

While research refers to the \ac{WER}, most commercial cloud providers that offer \ac{ASR} services avoid providing specific accuracy rates. Instead they use indeterminate claims like “hiqh-quality transcription” \cite{Microsoft2023}, “produce accurate transcripts” \cite{Amazon2023} or “state-of-the-art accuracy“ \cite{Google2023}. Government institutions and public organisations also remain vague in their requirements for the accuracy levels for captions. The \ac{FCC} requires captions to match the spoken words to the fullest extent possible, and additionally states that their rules on captioning distinguish between pre-recorded media and live transcriptions \cite{FCC2014}. The \ac{WCAG} only specify that captions are provided, without clearly defined requirements for accuracy \cite{WCAG2023}. However the W3C \ac{WAI} is a bit more specific by stating that \ac{ASR}-generated captions do not meet accessibility requirements unless they can be confirmed to be fully accurate \cite{WAI2023}. 

In order to progress in establishing accessibility standards, we need to make its accuracy more tangible. This paper provides a comprehensive and independent overview of the \ac{WER} for a variety of common \ac{ASR} services. The aim is to capture the accuracy of the current generation of \ac{ASR}-models. We do not promote specific vendors, as their individual models are subject to change. We created a novel dataset that attempts to reflect the performance of \ac{ASR} with minimal bias and without targeting specific weaknesses. To provide a high comparability of results, the transcription process was fully automated, using equivalent configurations across different vendors. The \ac{WER} was calculated after extensive text normalisation to reduce errors caused by non-semantic differences. Our results may support future research in choosing realistic error rates, particularly for qualitative evaluations and the development of accessibility metrics.

\section{Background}

The accuracy of \ac{ASR} rapidly increased with the development of machine learning, in particular through deep neural networks (DNN) \cite{PapersWithCode2023}. At the time of writing, most commercial services use hybrid systems that consist of multiple individual models (e.g. acoustic, language and lexical) \cite{Li2022}. In these hybrid models, the input of a model depends on the output of the previous model in the pipeline. Recently a new generation of \ac{ASR} systems have reported a further increase in accuracy through the use of \ac{E2E} models \cite{Baevski2020}. In contrast to hybrid models, a single objective function is used to optimise the whole network. \ac{E2E} training is especially interesting for academic research, as no expert knowledge of specific components like the acoustic model is required. This approach also offers some benefits for industry, as it simplifies the whole \ac{ASR} pipeline and it is easier to generate larger training data through self- or semi-supervised learning \cite{Zhang2022, Chung2022, Xu2021}.

\subsection{Transcription with ASR}

Manually creating accurate transcripts for spoken language is a laborious task, as speech is up to ten times faster than typing \cite{Wald2006A}. Transcription of real-time events is even more challenging, as there is no time to replay audio, and there is only a small window of time to correct texts. Thus, trained professionals are needed to accurately transcribe live events. A common method is \ac{CART}, where the transcript of a professional typist is presented live on a screen. To keep up with the speed of speech, special phonetic keyboards or stenography are used. An alternative method is "respeaking", in which professionals speak the speaker's words into a trained \ac{ASR} system and correct the resulting text using a keyboard. As both methods are very demanding, multiple professionals are required for longer events to give the operators rest periods. Besides the complexity of this task and high operational costs, a big issue is the limited availability of these professionals. At the same time, as more digital content is produced in the form of online presentations and remote meetings, the demand for transcription of events is increasing.

Fully automated transcription through \ac{ASR} appears to be an effective solution. YouTube creates automatic subtitles for videos, Zoom offers real-time captioning in meetings and PowerPoint enables automatic transcription during presentations. \ac{ASR} is rather cheap, has few limitations regarding availability or logistics, and requires no working breaks. Popular cloud providers and specialised companies offer \ac{ASR} as an on-demand service and multiple open-source models are publicly available. They support a wide range of languages, batch and real-time transcription or even adding custom vocabularies for a presentation, to detect infrequent words, acronyms or technical terms.

It is important to note that the results of \ac{ASR} and professional transcribers are quite different due to their inherent strengths and weaknesses \cite{Romero2009}. \Ac{AI} is good at keeping up with the speed of speech, and creates almost verbatim transcripts. The amount of words in an \ac{ASR}-generated transcript is almost identical to the amount of words spoken. But \ac{AI} can produce confusing words or sentences, by mistaking homonyms or hallucinating text. This can severely impair understanding, particularly for people who have to rely on captions. Humans on the other hand create less verbatim transcripts that concentrate on transporting the gist of the text, but even professional respeakers lag around 20-40 \ac{WPM} behind the actual speakers \cite{Romero2009}. Even though the transcript might be easier to read than results from \ac{ASR}, it can also miss important information. 

While more summarised captions may be selected in certain situations, such as children’s programs or when the speed of the caption presentation is very high, the majority of \ac{DHH} individuals tend to favour the comprehensive access offered by verbatim texts \cite{NIDCD2017}. The FCC also requires that offline captions must be verbatim, and paraphrasing should be minimised for real-time transcription \cite{FCC2014}. Besides text accuracy, accessible transcription incorporates other issues like the placement of captions, correct speaker identification or the presence of relevant non-speech information \cite{WAI2023, FCC2014}.

\subsection{Benchmarking ASR}

Evaluation of an \ac{ASR} system heavily depends on the use-case, and various factors have to be considered \cite{Aksenova2021}. A voice control application might focus on keyword spotting and correctly transcribing addresses or phone numbers. A transcription service for conversations and meetings, on the other hand, targets a high overall text accuracy. The speed of an \ac{ASR}-system is another critical factor. Real-time transcription aims for a short latency between the input audio stream and the text output. It is typically measured as the \ac{RTF} \cite{Liu2000}. Another factor is the stability of partial results, that can be transmitted before the final result in streaming \ac{ASR}. Volatile intermediate transcripts result in many text changes, which creates a negative user experience. There is no universal metric yet, but \citet{Shangguan2020} propose to measure the stability as an \ac{UPWR} and \citet{Baumann2009} suggest an Incremental \ac{WER}.

The quality of \ac{ASR} in poor acoustic environments has greatly improved, e.g. the accuracy in noisy environments \cite{Kinoshita2020}. But besides technical improvements in specific areas and the evolution of machine learning in general, the quality and amount of training data is the major factor influencing the performance of an \ac{ASR} system. Using more hiqh-quality training data increases the general accuracy of an \ac{ASR} system. While public datasets like the Switchboard corpus \cite{Godfrey1992} have been available since the 2000s, development is driven by additional and new datasets like LibriSpeech \cite{Panayotov2015}, and large multilingual audio recordings like CommonVoice \cite{Ardila2020}. 

The LibriSpeech corpus \cite{Panayotov2015} is a widely used benchmark in science and industry. With its release in 2015 the first models reported a \ac{WER} of 13.25\%. Within only six years, the most accurate models could decrease the \ac{WER} down to 2.5\% \cite{PapersWithCode2023}. However \citet{Geirhos2020} show that fine-tuning a model for a specific dataset carries the danger of shortcut learning, where a DNN exploits weaknesses in the training data. A model that reports excellent results on one speech corpus, might perform worse on other corpora. \citet{Chan2021} show that training a model on multiple public speech datasets results in a comparable accuracy but higher robustness than models trained on a single source. Respectively \cite{Gandhi2022} propose to evaluate \ac{ASR} systems on multiple public speech datasets resulting in an average \ac{ESB} score.

\subsection{Measuring transcription accuracy}

Research in speech recognition typically uses the \ac{WER} to measure the accuracy of \ac{ASR} systems. It is based on the minimum edit-distance between a transcript and the reference solution (aka ground truth), and quantifies the relative amount of errors to the total number of words of a text. But the \ac{WER} has been criticised, as it does not reflect text understanding, and only weakly correlates with human judgement of a transcripts’ quality \cite{Wang2003, Mishra2011, Favre2013}. Additional measures like the \ac{CER}, the \ac{MER} \cite{Morris2004} or the \ac{WWER} \cite{Apone2010} have the same underlying problem, as they also quantify the editing distance between two texts. Recent approaches try to measure the accuracy with \ac{AI}, as \citet{Kafle2017} initially proposed with the \ac{ACE} and extended with a second study \cite{Kafle2019}. But \citet{Wells2022} could not find a significant difference between \ac{ACE} and \ac{WER}. While \ac{AI} based metrics have the potential to better align with human judgement compared, they also bear the risk of a bias related to their training data. \ac{ACE} for example is trained on conversational speech and thus might not be suited for other scenarios like Higher Education or Entertainment.

Another metric, the NER-model was developed by \citet{Romero2015} to measure the accuracy of respeaking. It requires a manual analysis of the transcript in order to classify the errors by classifying their severity. As it focuses on respeaking, it also distinguishes between recognition and editing errors, to provide feedback for the respeaker. A drawback of the NER-model is that qualitative measures are subjective and hard to obtain on large datasets (e.g. training data for \ac{ASR}).

\begin{table*}
  \caption{Word Error Rate for different degrees of text normalisation.}
  \label{tab:normalisation}
  \begin{tabular}{lllr}
    \toprule
    Normalisation & Transcript & Sentence & WER \\
    \midrule
    None & Ground Truth & Hactar analysed, that the answer is not forty two! & - \\
    None & Hypothesis & \underline{hactar analyzed} that the answer \underline{isn't 42.} & 67\% \\
    \midrule
    Common & Ground Truth & hactar analysed that the answer is not forty two & - \\
    Common & Hypothesis & hactar \underline{analyzed} that the answer \underline{isnt 42} & 56\% \\
    \midrule
    Whisper & Ground Truth & hactar analyzed that the answer is not 42 & - \\
    Whisper & Hypothesis & hactar analyzed that the answer is not 42 & 0\%\\
    \bottomrule
  \end{tabular}
\end{table*}

Despite the criticism, the \ac{WER} is still useful to evaluate and compare \ac{ASR} systems. As these systems produce verbatim transcripts, a word by word comparison is suitable and an analysis in terms of the content becomes less relevant. The lower the \ac{WER} is, the closer the produced transcript is to the actual text - fewer errors should therefore result in high levels of understandability. However, the \ac{WER} is very vulnerable to non-semantic differences due to text formatting, for example errors resulting from incorrect capitalisation or punctuation. Table \ref{tab:normalisation} shows an example sentence and the resulting \ac{WER} for different text normalisations: None, Common (Removal of capitalisation and punctuation) and the Whisper normaliser.

To reduce the impact of these formatting differences, it is common to pre-process transcripts before the calculation of the \ac{WER}. Typically the text is transformed to lowercase and all punctuation is removed. \citet{Koenecke2020} introduce additional replacements for abbreviations and colloquial speech specific to the used dataset of their study. \citet{Whisper2022} developed an extensive normaliser for English, that unifies UK- and US-spelling differences, replaces common contractions and converts written numbers. Errors from capitalisation or grammatical cases are more common in languages other than English, but it is unclear whether these have an impact on the understandability of the text. The \ac{FCC} requires that capitalisation and punctuation is included in accurate captions, as these influence an individual’s understanding of the text \cite{FCC2014}. 

\subsection{Comparison of ASR services}

\begin{table*}
  \caption{Summary of Word Error Rate measurements of different services from related research}
  \label{tab:related-studies}
  \begin{tabular}{lcc|ccccccccccc}
    \toprule
    \multicolumn{3}{c}{\textbf{Source}} & \multicolumn{11}{c}{\textbf{WER of vendors included in this study}} \\
    & \rotatebox[origin=b]{90}{peer-reviewed} & \rotatebox[origin=b]{90}{public dataset} & \rotatebox[origin=b]{90}{Amazon} & \rotatebox[origin=c]{90}{AssemblyAI} & \rotatebox[origin=c]{90}{Deepgram} & \rotatebox[origin=c]{90}{Google} & \rotatebox[origin=c]{90}{IBM} & \rotatebox[origin=c]{90}{Microsoft} & \rotatebox[origin=c]{90}{Rev AI} & \rotatebox[origin=c]{90}{Speechmatics} & \rotatebox[origin=c]{90}{Speechtext.AI} & \rotatebox[origin=c]{90}{Tencent} & \rotatebox[origin=c]{90}{Whisper} \\
    \midrule
    
    \textbf{\cite{Addlesee2020}} & yes & yes & - & - & - & 7 & 6 & \textbf{5} & - & - & - & - & - \\
    \textbf{\cite{Ballenger2022}} & yes & no & 22 & - & - & 31 & 28 & \textbf{14} & - & - & - & - & - \\
    \textbf{\cite{Catania2019}} & yes & yes & - & - & - & 20 & \textbf{20} & - & - & - & - & - & -  \\
    \textbf{\cite{Koenecke2020}} & yes & yes & 13 & - & - & 14 & 16 & \textbf{10} & - & - & - & - & - \\
    \textbf{\cite{Tadimeti2022}} & yes & no & 18 & - & - & \textbf{11} & 27 & 16 & - & - & - & - & - \\
    \textbf{\cite{Tatman2017}} & yes & yes & - & - & - & \textbf{31} & - & 45 & - & - & - & - & -  \\
    \textbf{\cite{Chiu2018}} & yes & no & - & - & - & \underline{\textbf{6}} & - & - & - & - & - & - & - \\
    \textbf{\cite{Saon2017}} & yes & yes & - & - & - & - & \underline{\textbf{6}} & - & - & - & - & - & -  \\
    \textbf{\cite{Xiong2017}} & yes & yes & - & - & - & - & - & \underline{\textbf{5}} & - & - & - & - & -  \\
    \textbf{\cite{AssemblyAI2023}} & no & yes & - & \underline{\textbf{6}} & - & - & - & - & - & - & - & - & 7  \\
    \textbf{\cite{Deepgram2022}} & no & no & - & - & \underline{\textbf{11}} & - & - & - & - & - & - & - & 13  \\
    \textbf{\cite{RevAI2020}} & no & no & 18 & - & - & 16 & - & 17 & \underline{\textbf{14}} & 15 & - & - & -  \\
    \textbf{\cite{SpeechmaticsUrsa2023}} & no & yes & 17 & - & - & 19 & - & 15 & - & \underline{\textbf{12}} & - & - & 16  \\
    \textbf{\cite{Speechtext2023}} & no & yes & 16 & - & - & 9 & - & 14 & - & - & \underline{\textbf{4}} & - & - \\
    \textbf{\cite{Whisper2022}} & no & yes & - & - & - & - & - & - & - & - & - & - & \underline{\textbf{13}} \\
    \bottomrule
  \end{tabular}
\end{table*}

There is a lot of research that compares different \ac{ASR} services and focuses on bias regarding gender, race or age. \citet{Koenecke2020} report an almost twice as high \ac{WER} for Black American speakers compared to white American speakers. Despite the individual accuracy, these racial disparities exist across all tested vendors. They see the main problem in the performance gap of the acoustic model, due to insufficient training data featuring black American speakers. \citet{Tatman2017} also note that the accuracy of ASR systems depend on sociolinguistic factors, and is worse for Black Americans compared to white Americans. 

Obviously, \ac{ASR} does not distinguish between skin colours, but reflects biases in the training data. \ac{ASR} therefore performs worse for speakers of underrepresented groups within the training data. As a result the accuracy decreases for speakers with regional accents or second language learners. \citet{Tadimeti2022} report a performance gap between general American and non-American accents. This does not only apply to English, as \citet{Cumbal2021} show by comparing native and non-native speakers of Swedish. Bias in \ac{ASR} applies to all kinds of dimensions. \citet{Catania2019} show that the level of emotionality decreases accuracy compared to neutral speech. The company \citet{Speechmatics2023} reports that also the age of the speakers is a factor. \ac{ASR} shows the highest accuracy between the ages of 28 to 36, and the highest error rates for the oldest group of 60 to 81.

Other studies compare \ac{ASR} systems in general, without focusing on a specific bias. \citet{Addlesee2020} tested three common cloud providers with the Switchboard corpus and reported results between 5.1\% and 6.8\% \ac{WER}. \citet{Ballenger2022} transcribed lecture recordings of adult educational content on five common services and report a \ac{WER} ranging between 12\% to 31\%. Researchers from Microsoft report a \ac{WER} of 5.1\% on the Switchboard corpus \cite{Xiong2017}. Testing the same dataset, a research group from IBM reported a \ac{WER} of 5.6\% \cite{Saon2017}. These values are identical to \citet{Addlesee2020}’s results of Microsoft and IBM on the same corpus. Researchers from Google report a \ac{WER} of 5.6\% on an internal dataset of 12500 hours of Google Voice Search utterances \cite{Chiu2018}. In 2022 OpenAI released the open-source \ac{ASR} model Whisper. In the corresponding paper they show that this state-of-the-art \ac{E2E} model outperforms many commercial vendors on various public datasets \cite{Whisper2022}.

We also found data often part of advertising on company websites or whitepapers. While some vendors used public datasets to test the accuracy of their \ac{ASR} engines (SpeechText.AI \cite{Speechtext2023}, Rev AI \cite{RevAI2020}, Speechmatics \cite{SpeechmaticsUrsa2023}, AssemblyAI \cite{AssemblyAI2023}), others referred to internal datasets (Deepgram \cite{Deepgram2022}). When companies publish data on the accuracy of their \ac{ASR} services, it is not always clear whether these serve the purpose of advertising or a favourable comparison to other vendors. 

Table \ref{tab:related-studies} shows an overview of different studies, whitepapers and company websites that provide data on \ac{ASR} services. The lowest \ac{WER} value (best result) of each row is highlighted bold. Results where study authors and lowest \ac{WER} are from the same company are underlined. The values of each study are not comparable to others, as the error rates depend on multiple factors: (1) the tested dataset, (2) the date of the test, and (3) the degree of text normalisation to calculate the \ac{WER}.

In 15 different tests, nine different providers delivered the best result. At maximum five different vendors are compared within a study or test. While the common cloud providers Amazon, Google, IBM and Microsoft appear in multiple studies, specialised \ac{ASR} providers are less represented. The error rates have a very large range from 3.8\% to 45.0\%, which most likely results from the different datasets used and the degree of text normalisation.

\section{Methodology}

\begin{figure}[t]
  \centering
  \includegraphics[width=\linewidth]{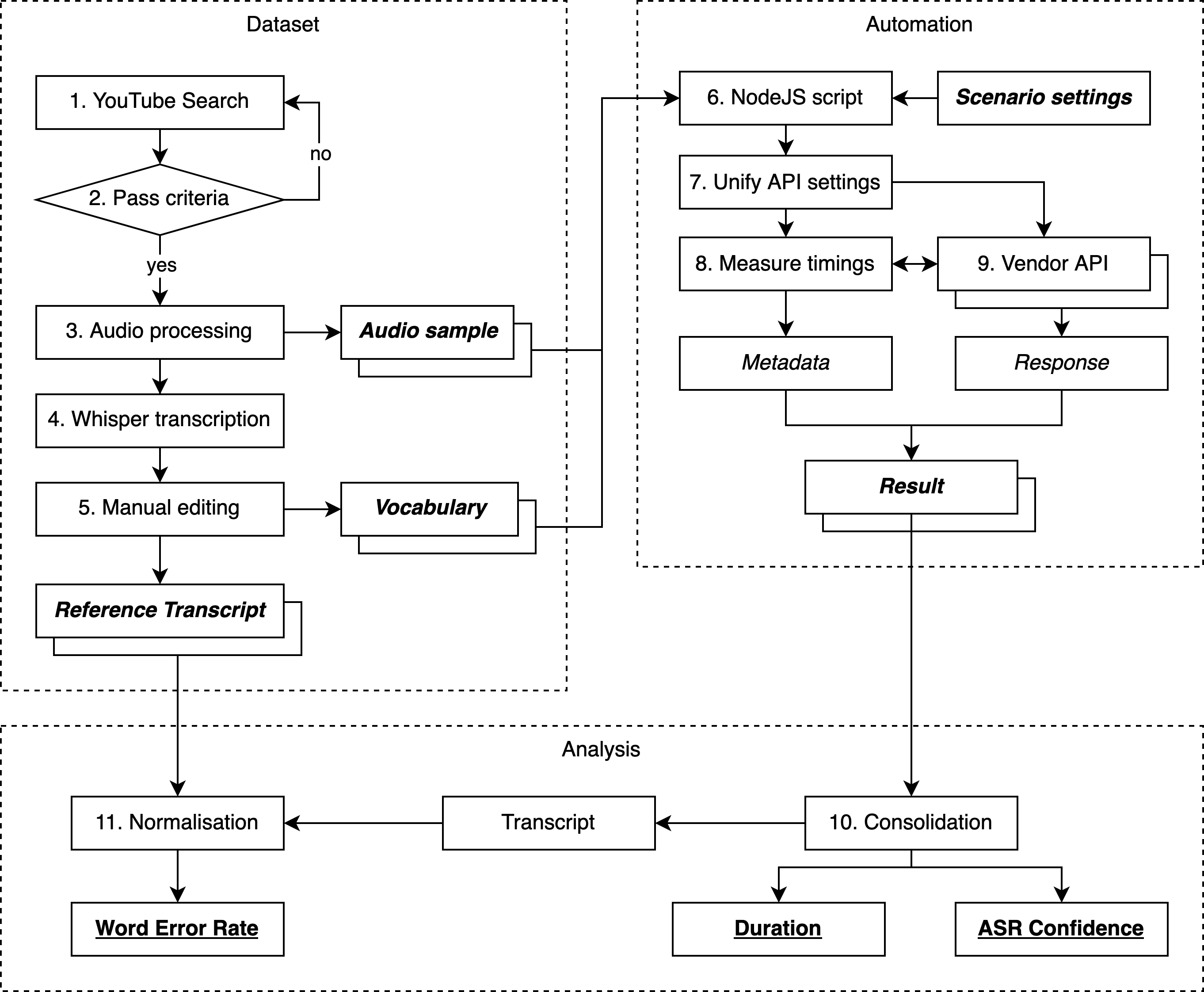}
  \caption{Methodology}
  \Description{A UML diagram that visualises the methodology of the study. The diagram shows three areas: the curation of the dataset, the automation of the transcription, and the analysis. The individual steps of the diagram are briefly referenced at the beginning of the methodology section and explained in detail in its subsections.}
  \label{fig:methodology}
\end{figure}

The methodology is illustrated in Fig. \ref{fig:methodology} and explained in detail in the following sections. The dataset contains lecture recordings obtained through a systematic YouTube search (1). The search results were filtered according to a set of selection criteria (2) to ensure conformity with the dataset. A randomly selected segment of the entire audio clip was extracted and pre-processed into a standardised audio format (3). After an initial transcription using Whisper (4), the transcript was manually corrected (5) to create an error-free reference and a vocabulary list.

A NodeJS script automates transcription across different \ac{ASR} service providers (6). The dataset and scenario settings (batch or stream transcription; with or without vocabulary) are used as input for execution. The script unifies the \ac{API} settings (7), measures the duration of the transcription process (8) and calls the vendor \ac{API}s (9). The response from the \ac{API} and additional metadata are stored as the result.

For the analysis, the different \ac{API} response formats were consolidated (10). Prior to calculating the \ac{WER}, the \ac{ASR} transcript and the reference transcript were normalised. The duration and \ac{ASR} confidence were calculated from the stored metadata.

We evaluated eleven different \ac{ASR} services with our dataset, that contains 120 audio samples with a duration of 3 minutes (90 YouTube recordings; 30 control samples from the LibriSpeech test-other corpus). Each service processed between 180 and 480 transcription jobs, depending on supported languages and availability of streaming \ac{ASR} (see Table \ref{tab:vendor-features}). In total, 221 hours of audio were transcribed, resulting in 3840 individual transcriptions.

\subsection{Dataset}

The goal of this study was to capture the average performance of \ac{ASR} under fairly ideal but realistic conditions, rather than to identify specific weaknesses or biases. To achieve this objective, a new dataset was created to minimise the risk of samples being part of an \ac{ASR} model's training data, which could lead to a particularly good performance.

\subsubsection{Recording scenario}

Recordings were made in a Higher Education setting and consisted of (undergraduate) lectures or invited talks. We chose Higher Education as our use case as there is a high demand for affordable and scalable transcription of lectures, talks and other learning formats. Higher Education institutions are required to offer equal access to all students, and the transcription of lectures is an essential part of that, benefiting not only \ac{DHH} students but also (second) language learners and students taking part remotely. The scenario of Higher Education is also suitable, as it contains specialist vocabulary and lecturers are used to presenting without being professional speakers. We only chose parts of recordings, where only the presenter speaks, to avoid additional issues like speaker diarisation.

\subsubsection{Selection criteria}

We selected only videos published under a Creative Commons License from YouTube, which reduced the possible sample size significantly. We used keywords like "university lecture", "invited talk" and corresponding search terms for German videos. We excluded any videos that covered unsuitable topics (e.g. lectures not recorded in the context of Higher Education, talks by religious groups, interest groups or political parties). We selected only videos with an average audio quality, that (we think) is representative of a Higher Education setting (e.g. speakers using a clip-on microphone). We avoided recordings with either very professional setups, or bad audio quality (e.g. strong reverberation, noise interferences, unintelligible or barely audible speech). We additionally measured the quality of all speech recordings with MOSNet and selected only recordings with an absolute value of 2 or higher \cite{Lo2019}. We also excluded all videos that had manually created or corrected captions, as these might be part of the training data of Whisper and other services.

This initial selection of videos was then rated in a two step review process by three members of the team with regard to the following criteria: 

\begin{itemize}
\item{recording quality}
\item{for English samples: perceived difference from \ac{RP} in the UK or \ac{GenAm} in the USA}
\item{for \ac{ESL} samples: perceived intelligibility of speech}
\item{for German samples: High German without strong accents}
\end{itemize}

We used these criteria because heavily accented speech is challenging for \ac{ASR} systems, as it contains acoustic and linguistic markers of both the source and target language, and could thus be the source of transcription errors. Similarly, \ac{ASR} engines are often trained on datasets with \ac{RP} or \ac{GenAm}. Samples with a large difference from this training set, could produce more errors. It is a general problem that \ac{ASR} engines are only trained on limited datasets that do not include other variants of English as this could lead to or enforce existing biases.

\subsubsection{Bias}

The scenario of Higher Education leads to a bias in the dataset. For example, there is a higher number of recordings with white male speakers compared to Black female speakers. There are also many videos on some topics like Computer Science and few on other topics like Linguistics. We try to limit this bias by selecting content that covered a wide area of topics from Computer Science, Electrical Engineering and Chemistry to History, Sociology, Politics, Linguistics, Religion and Neuroscience. We selected samples with \ac{ESL} speakers who had different accents (European, South American, Asian). However, we did not select speakers for the \ac{ESL} samples from countries or regions that might have English as an official government language (e.g. India or Nigeria), as they might be more fluent or completed their education in English. The dataset has an even distribution of male and female speakers, but is not specifically designed to examine gender differences in \ac{ASR} accuracy.

\subsubsection{Data preparation}

We randomly selected a part of each video that had continuous and fluent speech (e.g. no interactions with the audience or writing on a board). The target length of the sample was three minutes. We avoided cutting the recording sample during a sentence or a word. The samples have an average duration of 184 seconds (min 174, max 209). The number of spoken words is 455 on average (min 287, max 671). The average \ac{WPM} is 148 (min 94, max 222). The audio was converted to a standardised format that all vendors supported (44,100 Hz, mono, PCM 16-bit, LE). Additionally, the volume of the audio was normalised with sox \cite{SOX2023}.

\subsubsection{Availability}

Even though we consider our study to meet the Fair Use\footnote{https://www.copyright.gov/fair-use/index.html} principle, and all recordings are publicly available videos under the Creative Commons licence, we do not want to publish our dataset without the individual speakers’ consent. We provide a detailed table on GitHub with the used recordings, time ranges, speaker origins and topic of the presentations.\footnote{https://github.com/shuffle-project/asr-comparison} To contextualise the results of our recordings we included a public dataset that many vendors or research studies use for training, model validation and comparison to other \ac{ASR}-models. We randomly selected 30 samples from the most common English corpus LibriSpeech test-other \cite{Panayotov2015} as a control dataset. The results also help to identify if \ac{ASR}-services perform particularly well on LibriSpeech. The recordings fulfill the same criteria as the English videos with regard to speaker accents and audio length.

\subsection{Additional variables}

Besides each samples’ individual speaker and content, this study uses additional variables to represent a wider range of material and to cover potential sources of bias in ASR systems. We chose 15 male and female speakers per language group. We have four language groups: English, \ac{ESL}, German and LibriSpeech (English). 

Additionally, we have four different technical scenarios: batch or stream transcription and with or without an additional custom vocabulary. Some vendors did not provide all technical requirements e.g. custom vocabularies, streaming or specific combinations (e.g. German streaming with a custom vocabulary). Table \ref{tab:vendor-features} lists the supported features of the vendors at the date of the test.

\begin{table*}
  \caption{Supported features by service provider (en = English, de = German)}
  \label{tab:vendor-features}
  \begin{tabular}{lcccc}
    \toprule
    Vendor & Batch & Streaming & Batch with vocabulary & Streaming with vocabulary \\
    \midrule
    
    \textbf{Amazon} & en, de & en, de & en, de & en, de \\
    \textbf{AssemblyAI} & en, de & \textbf{en} & en, de & \textbf{en} \\
    \textbf{Deepgram} & en, de & en, de & en, de & en, de \\
    \textbf{Google} & en, de & en, de & en, de & en, de \\
    \textbf{IBM} & en, de & en, de & \textbf{-} & \textbf{-} \\
    \textbf{Microsoft} & en, de & en, de & en, de & en, de \\
    \textbf{RevAI} & en, de & en, de & en, de & \textbf{en} \\
    \textbf{Speechmatics} & en, de & \textbf{en} & en, de & \textbf{en} \\
    \textbf{SpeechText.AI} & en, de & \textbf{-} & \textbf{-} & \textbf{-} \\
    \textbf{Tencent} & \textbf{en} & \textbf{-} & \textbf{en} & \textbf{-} \\
    \textbf{Whisper} & en, de & \textbf{-} & \textbf{-} & \textbf{-} \\

    \bottomrule
  \end{tabular}
\end{table*}

\subsection{Transcription}

Previous research suggests using \ac{ASR} as a basis for manual corrections to create more accurate transcripts compared to a fully manual approach \cite{Che2017}. We could not rule out the possibility that a repeated transcription of the same audio file influences the result of the second run (e.g. faster processing time through caching), except for Whisper. We therefore initially transcribed all samples with Whisper. Two independent coders manually corrected each transcript. We focused on creating a verbatim transcript and besides correcting mistakes, we carefully looked out for Whisper specific errors like the removal of duplicate words. The manual transcription included additional research to find the correct spelling of names and technical terms.

We encountered several problems during transcription, such as differences between UK- and US-spelling, common contractions (e.g. "it’s", "can’t") or currency writings (e.g. \$ 12000). Also for German we found colloquial spellings (e.g. "en" for "ein", "gsagt" for "gesagt"). Additionally, transcribing correct punctuation is difficult for spoken language and influences capitalization, e.g. by ending a sentence after a break or separating it with a comma. We therefore use the extensive text normalisation from Whisper and added additional replacements for specific words occurring in our German transcripts.

\subsection{Vendors}

We used the Google Search Engine to find the most relevant vendors that provide an \ac{API} for fully automated transcription of audio or video files. We excluded services that focus on specific tasks like keyword spotting or telephone transcription. We selected ten commercial vendors (Amazon AWS, AssemblyAI, Deepgram, Google Cloud Platform, IBM Watson, Microsoft Azure, Rev AI, Speechmatics, SpeechText.AI, Tencent Cloud) and OpenAI Whisper, as a self-hosted open source alternative. To minimise the effects of audio processing that vendors might use, we converted all samples to a standard audio format and normalised the volume.

We checked all \ac{API} parameters of each vendor to assure comparability and to avoid unintentional default settings. If vendors provided multiple models, we selected the model that promised the highest accuracy. These enhanced models might influence the transcription speed and costs (e.g. higher processing power). We did not use models for specific use-cases like transcription of medical recordings. To receive verbatim transcripts, we disabled all filters (e.g. profanity). We enabled detailed results if possible, including punctuation, text formatting, word level timestamps and confidences. With regard to Whisper, we used the large-v2 model and disabled “condition\_on\_previous\_text”, as it hallucinated and created longer transcripts than the actual file.

Some \ac{API}s can return multiple results and provide alternative transcriptions, sorted by the \ac{AI}’s confidence in the text correctness. We limited the results to one, to avoid additional processing time and analysed the text that the AI categorises as best fit. For stream processing we enabled partial results, to receive intermediate and final transcripts.

\subsection{Automation}\label{sec:automation}

We used NodeJS to automate all transcription jobs. If possible, we used the official \ac{SDK} of a vendor. Otherwise we implemented the HTTP- or WebSocket-API according to the vendors’ documentation. The streaming \ac{API}s utilise WebSocket connections that expect raw audio data either as an integer buffer or base64-encoded. A file stream splits the audio into 10 KB chunks and these parts are sent sequentially. The results might differ from an actual real-time stream, as the data are transmitted faster. However, in our test we did not find any indications that an \ac{API} buffered all data and processed the audio as whole. In contrast to all other vendors’ streaming \ac{API}s, Tencent does not provide a WebSocket-API. We therefore excluded Tencent from streaming transcription, as we see a possible issue in comparability due to the implementation.

In addition to the \ac{API} responses itself, we measured timings locally to calculate specific durations (e.g. upload, vocabulary creation, transcription, download). Some vendors require additional steps like uploading files to a cloud bucket, while others handle multiple steps within a single HTTP-Call (e.g. transcription with a custom vocabulary). The different approaches might affect the measured timings and are considered for a comparative analysis. We always used the closest region to our location (e.g. eu-central), if cloud vendors offered multiple regions.

The cloud transcriptions were run on 22.06.2023. In a few cases, we had to restart a transcription job, as an HTTP or API error occurred. We had no issues after a second try. The Whisper transcription was executed on the same day on a Macbook Pro 2020 with 16GB Ram and an Apple M1 ARM processor.

\subsection{Vocabularies}

Some \ac{API}s offer the functionality to upload custom vocabularies before the actual transcription. These vocabularies are lists of words or phrase sets. These words will be weighted higher by the AI and prioritised during transcription, especially if the \ac{AI} has similar probability values for multiple words. We created a vocabulary for each transcript consisting of abbreviations and names of persons, countries, regions, cities, rivers and buildings. We focused on an approach that is feasible in a Higher Education setting, e.g. automatically analysing the content of slides and filtering rare words. Therefore we did not add “sounds alike” terms that provide additional information to the \ac{AI} (e.g. how the phrases are pronounced). In the case of Tencent, we had to exclude all words that exceeded the limit of ten characters.

\subsection{Analysis}

We used the jiwer \cite{Jiwer2023} Python implementation to calculate the \ac{WER}, and applied extensive text transformation to all transcripts to reduce the impact of formatting differences on the error rate. We used the Whisper text normaliser \cite{Whisper2022}, which transforms text to lowercase, removes all punctuation, and unifies common contractions, spelling differences and spoken numbers. We extended the normalisation for German according to common contractions and numbers appearing in the transcripts of our dataset. Consequently as punctuation was removed for the calculation of the \ac{WER}, we only used confidence outputs from an \ac{API} on words and excluded punctuation marks.

To examine the effect of custom vocabularies, we counted the number of vocabulary hits. Every occurrence of a word from the vocabulary in a transcript is defined as a hit. The text normalisation was also applied to these words. We also measured the time taken for various processing steps like uploading, vocabulary creation, transcription and downloading of the results as described in section \ref{sec:automation}.

\section{Results}

\subsection{Accuracy}

\begin{table*}
    \caption{Average Word Error Rate in percent by service provider and dataset}
    \label{tab:accuracy}
    \begin{tabular}{lcccc}
    \toprule
    Vendor & English & LibriSpeech & ESL & German \\
    \midrule
    Amazon & 4.4 & 6.1 & 7.1 & 18.0 \\
    AssemblyAI & 4.5 & 4.2 & 5.9 & 13.2 \\
    Deepgram & 8.3 & 12.8 & 11.4 & 19.3 \\
    Google & 20.1 & 23.6 & 28.1 & 18.1 \\
    IBM & 11.2 & 13.2 & 17.3 & 20.6 \\
    Microsoft & 4.4 & 5.9 & 6.7 & 10.1 \\
    Rev AI & 4.4 & 7.0 & 6.7 & 19.2 \\
    SpeechText.AI & 8.4 & 11.3 & 14.1 & 21.4 \\
    Speechmatics & 3.3 & 3.6 & 4.6 & 8.0 \\
    Tencent & 4.6 & 4.1 & 7.3 & - \\
    Whisper (large-v2) & \textbf{2.9} & \textbf{3.3} & \textbf{3.3} & \textbf{5.0} \\
    \midrule
    Average & 7.0 & 8.6 & 10.2 & 15.3 \\
    \bottomrule
    \end{tabular}
\end{table*}

Table \ref{tab:accuracy} shows the average vendors’ \ac{WER} on the four different datasets for batch transcription without vocabulary. On average native English speakers achieved the lowest error rates and \ac{ESL} speakers performed better compared to the results for German Native Speakers. Comparing the results of English and the LibriSpeech dataset most vendors show a comparable quality. Deepgram, Google, Rev AI and SpeechText.AI had considerably higher error rates for LibriSpeech. The lower results of AssemblyAI and Tencent could indicate an optimisation towards the LibriSpeech corpus. Interestingly, Google was the only vendor where German performed best compared to all other datasets, most likely due to the comparatively poor performance of the English \ac{ASR} model.

\begin{figure}[t]
  \centering
  \includegraphics[width=\linewidth]{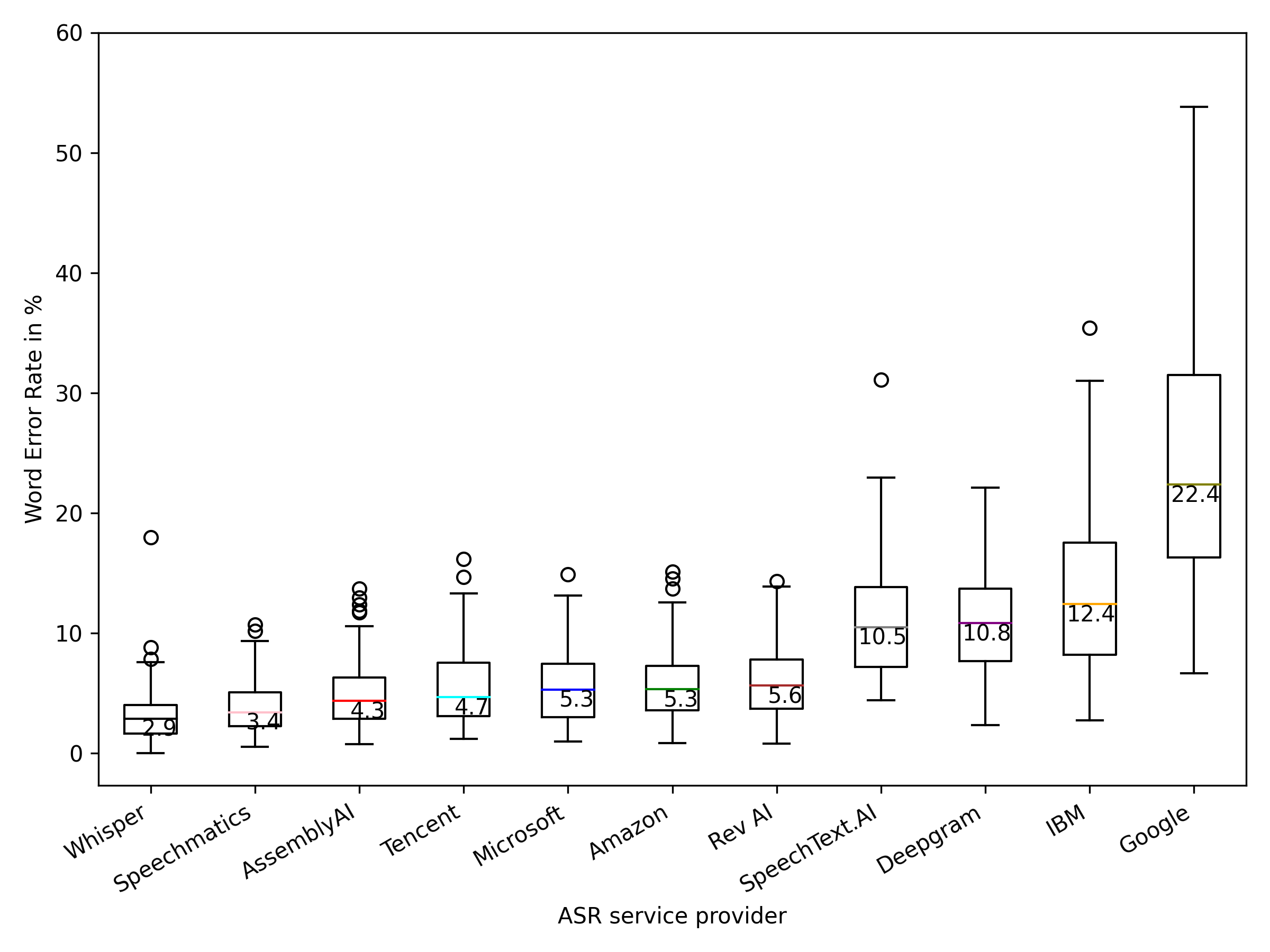}
  \caption{Word Error Rate by vendor for the English datasets}
  \Description{Boxplots of the Word Error Rate in percent for the eleven service providers. The mean values range from 2.9 to 22.4. Whisper and Speechmatics have the lowest mean values and low standard deviations. AssemblyAI, Tencent, Microsoft, Amazon and Rev AI have a mean value around 5. SpeechText.AI, Deepgram and IBM around 10. Google has the highest mean and standard deviation. In general the providers have only a few outliers.}
  \label{fig:accuracy-by-vendor}
\end{figure}

Fig. \ref{fig:accuracy-by-vendor} shows the \ac{WER} by vendor for all English datasets transcribed as batch and without vocabulary. The average \ac{WER} across all vendors was 7.0\%. The lowest calculated value was 0\% and the highest 53.8\%. The results show that both the average accuracy and standard deviation of error rates varies between the vendors. Service providers with a high average accuracy tend to have a smaller standard deviation compared to providers with a low average accuracy.

\begin{figure*}[t]
  \centering
  \includegraphics[width=\linewidth]{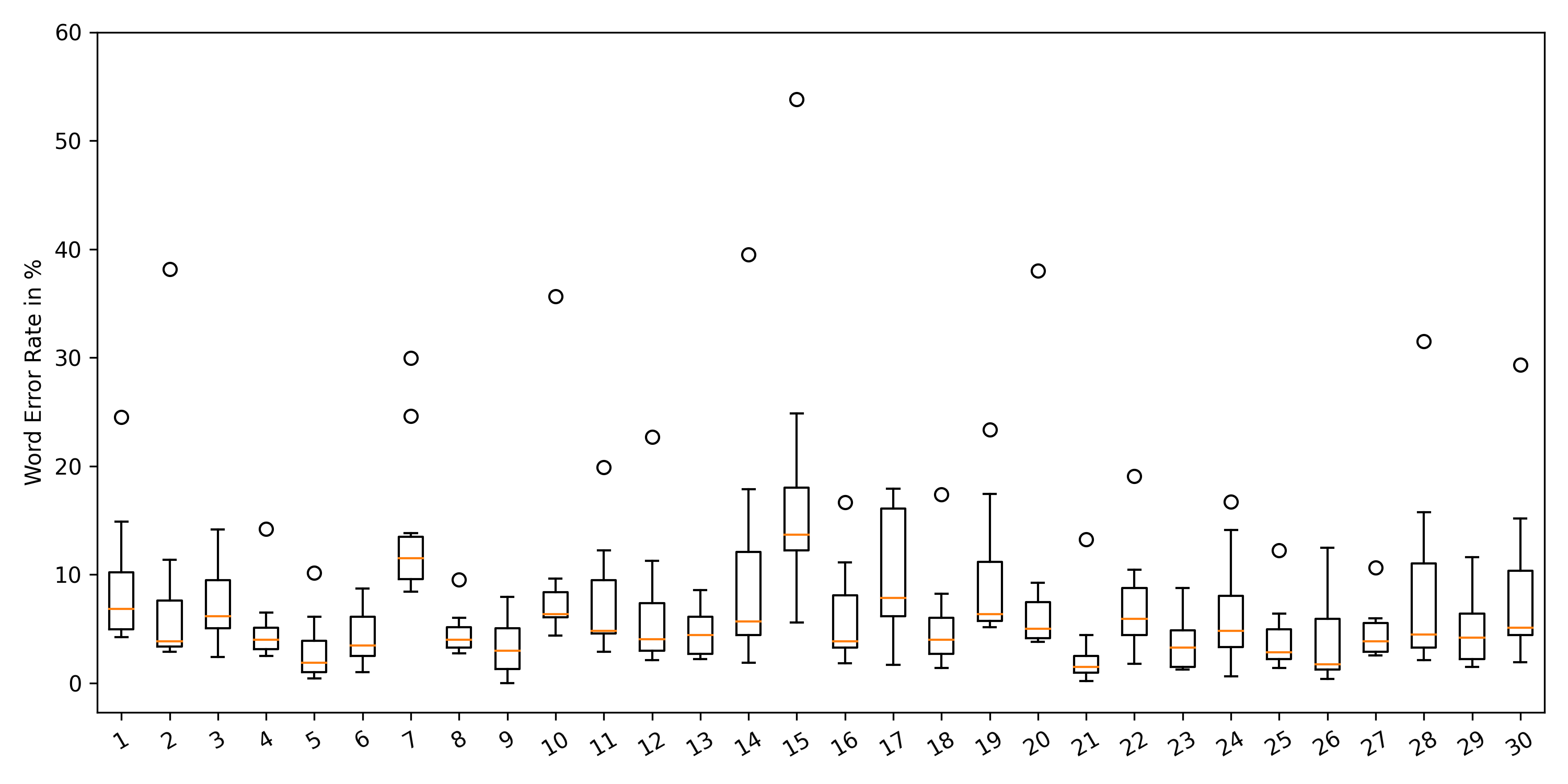}
  \caption{Word Error Rate by file for the English datasets}
  \Description{Boxplots of Word Error Rate in percent for the 30 English audio samples. The smallest distribution within an audio sample is around 3\%, while the highest is around 20\%. Samples with lower means show smaller standard deviations compared to samples with higher means. In general samples have at maximum one outlier.}
  \label{fig:accuracy-by-file}
\end{figure*}

Fig. \ref{fig:accuracy-by-file} shows the average \ac{WER} per audio sample. The results show that mean and standard deviation varies between the individual samples. Samples with a high average accuracy tend to have a smaller standard deviation compared to samples with a low average accuracy. For the 30 different samples, we counted seven different vendors that achieved the highest accuracy and three different vendors for the lowest accuracy.

\subsection{Vocabulary}

Eight vendors offered the functionality to provide a custom vocabulary for words or phrases that might appear in the audio. The average \ac{WER} of these providers across the English datasets was 8.61\%. By providing a vocabulary, it was reduced to 8.10\%. We performed a Welch’s t-test to investigate whether the \ac{WER} differed between transcriptions with and without an additional vocabulary. There was no significant difference in the \ac{WER} of the transcripts without vocabulary (M=0.086, SE=0.003) and with a vocabulary (M=0.081, SE=0.003) speakers (t (1438)=-1.646, p=0.640). The effect size was moderate (r=0.580).

\begin{figure}[t]
  \centering
  \includegraphics[width=0.8\linewidth]{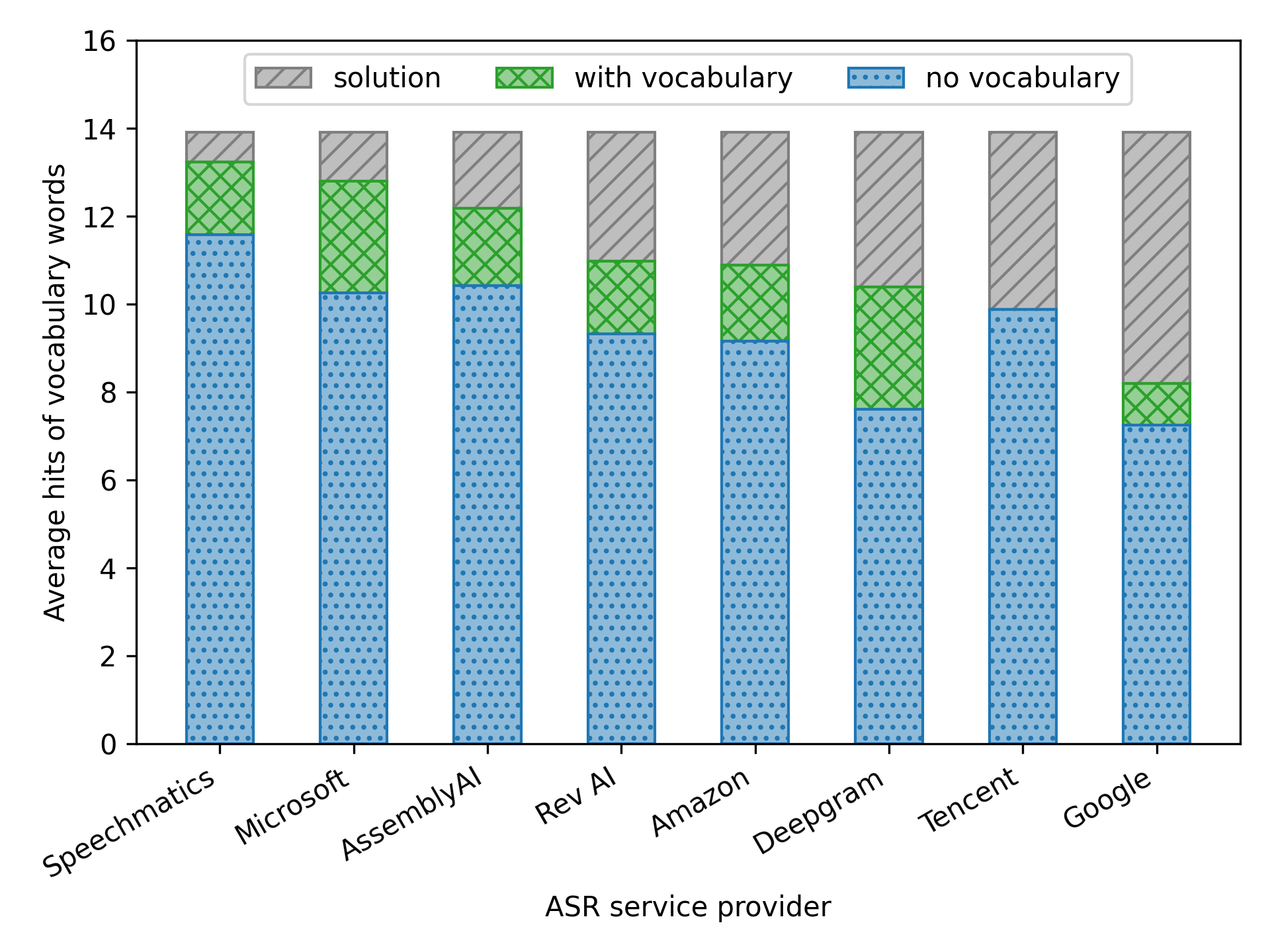}
  \caption{Average hits of vocabulary words by vendor for the English datasets}
  \Description{Stacked bar chart showing average vocabulary hits from 0 to 16 on the Y axis by service provider on the X axis. Each bar is a stack with the condition "no vocabulary" on the bottom, "with vocabulary" in the middle and "solution" on the top. Bars representing hits with no vocabulary are the largest. They are distributed from around 8 to 12 hits. Bars showing hits with vocabulary are smaller and between 0 and 3. Bars showing the hits in the solution transcript range from around 1 to 6.}
  \label{fig:vocabulary-hits}
\end{figure}

Fig. \ref{fig:vocabulary-hits} displays the average hits of these vocabulary words within the transcripts for each service provider. For all vendors the majority of hits was already counted without a vocabulary. The hits increased for all vendors except Tencent when providing a vocabulary. No vendor reached the maximum hits according to the reference transcript.

\subsection{Streaming}

On the English datasets streaming transcription had a higher WER (10.9\%) compared to batch transcription (9.37\%). We performed a Welch’s t-test to investigate whether the \ac{WER} differed between batch and streaming transcription. There was a significant difference in the \ac{WER} of the transcripts for batch (M=0.094, SE=0.003) and streaming (M=0.109, SE=0.003) transcription (t (1438)=-1.646, p<0.01). The effect size was large (r=0.995).

\subsection{Confidence}

\begin{figure}[t]
  \centering
  \includegraphics[width=0.8\linewidth]{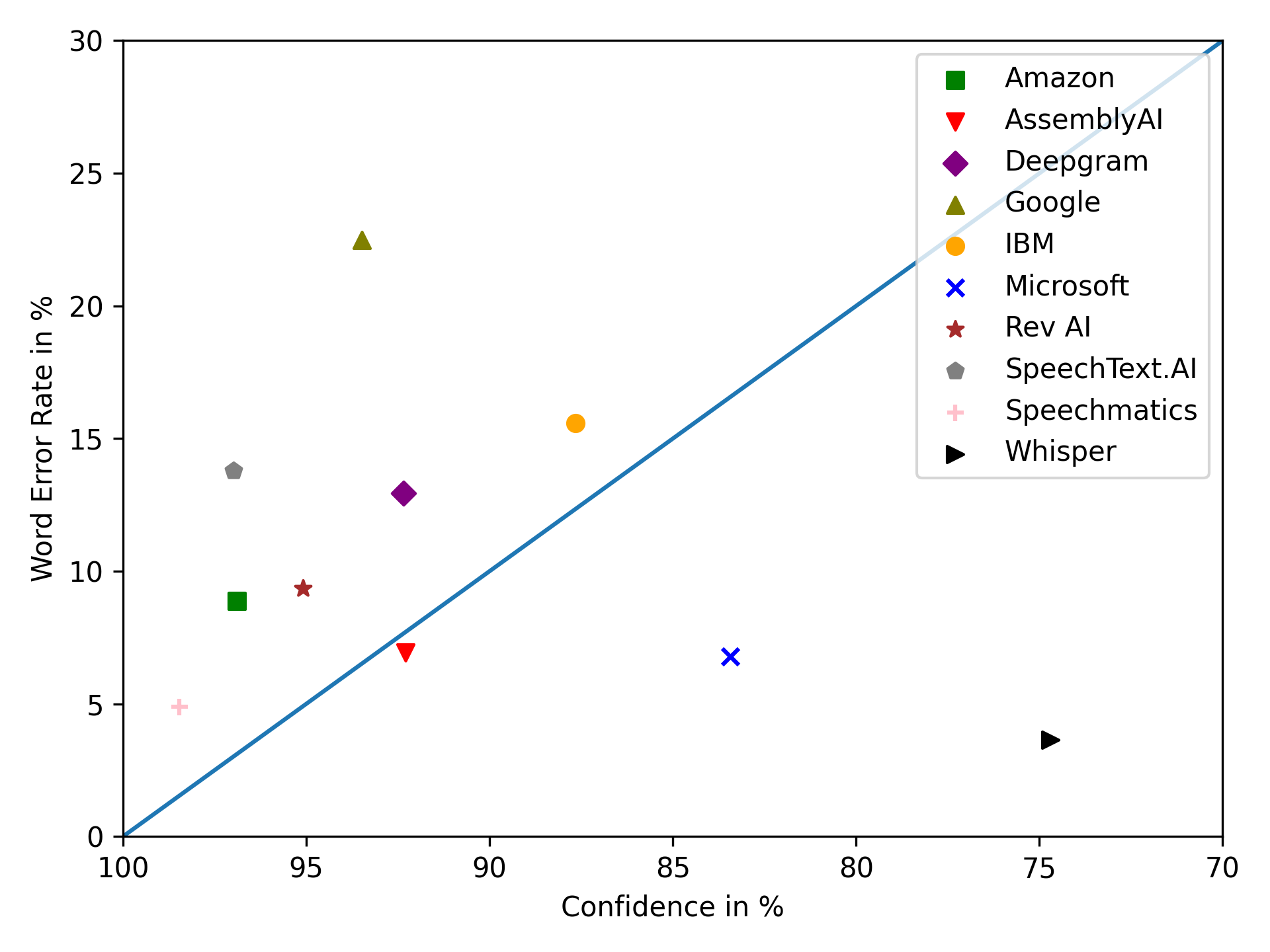}
  \caption{Average Word Error Rate in percent to confidence in percent by vendor for all datasets}
  \Description{Scatter plot showing the vendors' average Word Error Rate in percent on the Y axis from 0 to 30 and the average confidence in percent on the X axis from 100 to 70. A blue diagonal line starts in the bottom left to the top right. Most points are not positioned near the blue diagonal line. The majority of points are above the diagonal line in the left half of the diagram. AssemblyAI is almost on the blue diagonal line. Microsoft and Whisper are below the diagonal line in the right half of the diagram.}
  \label{fig:wer-confidence}
\end{figure}

Fig. \ref{fig:wer-confidence} shows the average confidence by \ac{WER} for each vendor. Tencent did not provide confidence values. The results from batch transcription for all datasets are used. Ideally the confidence aligns with the \ac{WER}. If an AI is 80\% confident, the measured \ac{WER} should be 20\%. The blue diagonal line represents this ideal relation between confidence and \ac{WER}. Data points in the upper left half represent overconfidence and data points in the lower right half underconfidence. Google and SpeechText.AI tend to have higher confidences than the measured accuracy, while Microsoft and Whisper report lower confidences. The other services show a higher alignment between the measured \ac{WER} and confidence output. Only AssemblyAI shows an ideal relation between \ac{WER} and confidence on average. Considering the varying deviations of the other providers, this could be a mere coincidence.

\subsection{Duration}

\begin{figure}[t]
  \centering
  \includegraphics[width=0.8\linewidth]{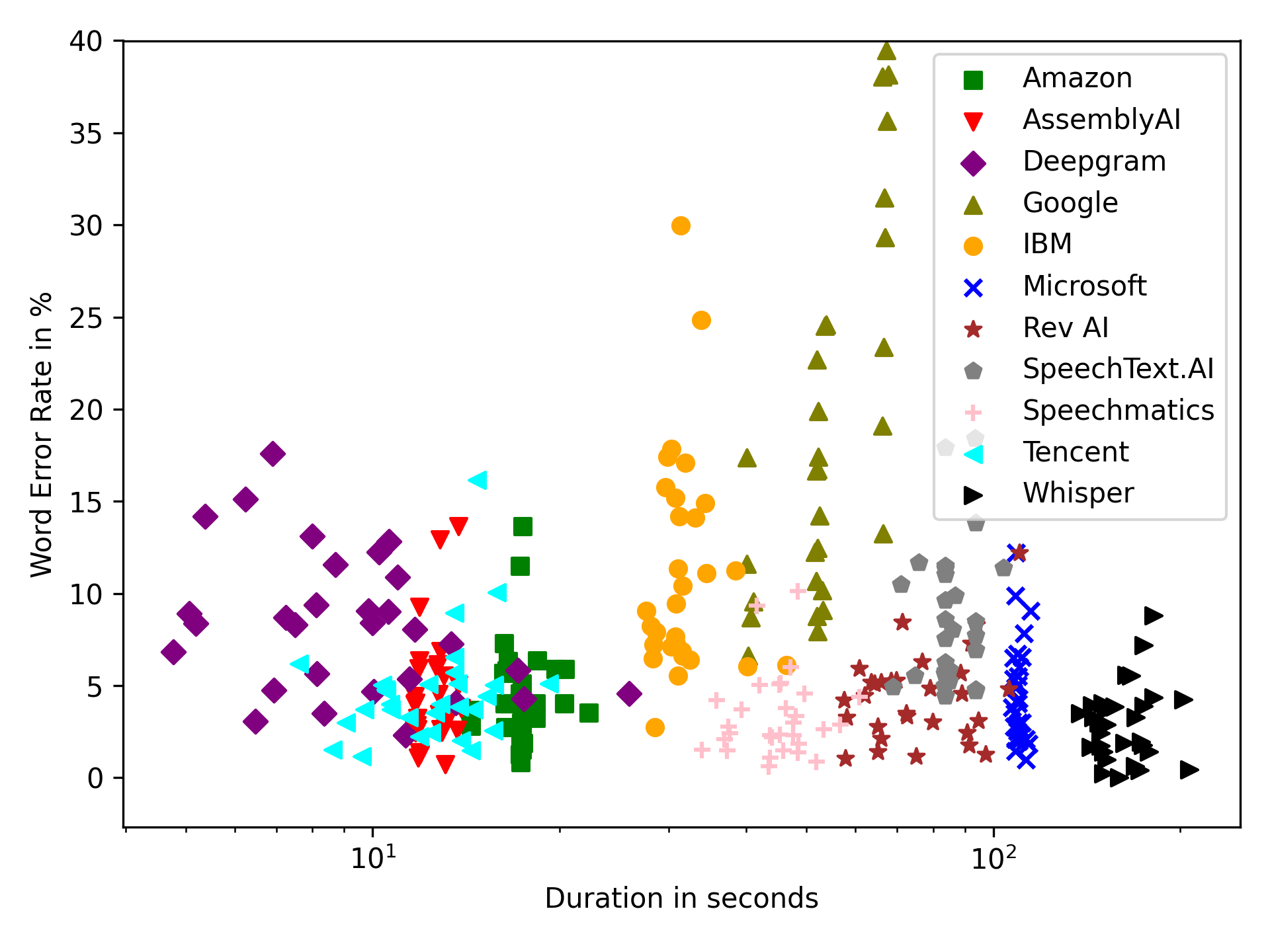}
  \caption{Word Error Rate in percent to processing duration in seconds by vendor for the English dataset}
  \Description{Scatter plot showing the Word Error Rate in percent on the Y axis from 0 to 40 and the processing duration in seconds on the X axis in a logarithmic scale from 0 to 400. Most points have a WER lower than 20 and a duration lower than 150. Only a few points have a WER over 30 or duration over 200.}
  \label{fig:wer-duration}
\end{figure}

The duration an \ac{ASR} service requires to process a file depends on multiple factors, like the complexity of the \ac{ASR} pipeline, the model size and mainly the computing power that is provided by the vendor. It can also be reduced, by splitting the audio into small chunks to process them in parallel. Fig. \ref{fig:wer-duration} displays the \ac{WER} of the transcript to the processing duration for each transcription job of the English dataset. Most vendors show a fixed duration independent of the resulting accuracy. This is as expected, as all audio samples are around three minutes long. Vendors that have a longer processing duration do not show lower error rates. Whisper has the highest accuracy and longest processing duration, which can either result from the model complexity or the fact that it was executed on consumer hardware.

\subsection{Text normalisation}

The degree of text normalisation strongly impacts the calculated \ac{WER}. Without any transformation the average \ac{WER} of all datasets was 21.83\%. Transforming the text to lowercase reduced the \ac{WER} to 18.15\% and removing punctuation to 15.48\%. Both normalisations combined resulted in an average \ac{WER} of 11.29\%. The additional replacements introduced by the Whisper normaliser, and our extension for German reduced the average \ac{WER} to 10.16\%. 

We performed a Welch’s t-test to investigate whether the \ac{WER} differed between the simple normalisation (lowercase and no punctuation) and the extended normalisation (Whisper normaliser and our German additions). There was a significant difference in the \ac{WER} of the transcripts for simple (M=0.113, SE=0.002) and extended (M=0.102, SE=0.002) normalisation (t (2578)=-1.645, p<0.01). The effect size was large (r=0.977).

\subsection{Alternative error metrics}
The \acl{MER} (M=0.099, SE=0.080) shows similar results compared to the \ac{WER} (M=0.102, SE=0.083). While the average and standard deviation of the \acl{CER} (M=0.051, SE=0.050) are around 50\% lower, they are around 30\% higher for the \acl{WIL} (M=0.152, SE=0.118).

\section{Discussion}

Our study evaluated eleven common \ac{ASR} services by transcribing recordings from Higher Education lectures. Results show that the accuracy varied strongly between the different vendors. Even providers that achieve a relatively low average \ac{WER}, can show a high error rate for an individual audio sample. Among the quality differences between vendors, this volatility is also shown in the accuracy distribution for each recording. Even though the samples do not contain strong accents, the performance of \ac{ASR} heavily depends on the individual speaker and acoustic environment. Different vendors provided the highest or lowest scores for each recording. There was not one vendor who consistently reached the lowest \ac{WER} across all samples. The variance in \ac{WER} between the individual recordings show that it is hard to come up with one reliable number to score the accuracy of an \ac{ASR} provider, even for a quite homogeneous dataset like ours. For English the state-of-the-art average value seems to be around 5\%, which aligns with the reports of related studies \cite{Addlesee2020, Chiu2018, Saon2017, Xiong2017, AssemblyAI2023}. However, our findings do not necessarily disprove other studies that report higher error rates (see Table \ref{tab:related-studies}), as the \ac{WER} is heavily dependent on the degree of text normalisation.

As our samples are recordings in the context of Higher Education, results might be different for more spontaneous, conversational, and colloquial speech. However, insufficient accuracy of \ac{ASR} is the main issue reported by \ac{DHH} individuals in related studies \cite{Kawas2016, Butler2019} and by the DHH community \cite{NADPetition2019, NDC2020}. This makes it difficult to assess \ac{ASR}’s ability to provide accessible transcripts, as it depends on a case-by-case basis.

Apart from the different services and individual speakers, language itself has the biggest impact on the transcription’s accuracy. \ac{ASR} achieves the lowest error rates for native English speakers. While the samples of non-native speakers show higher error rates for English, they still perform better compared to our German dataset. The quality most likely decreases for languages with a smaller corpus of training data, and more complex languages, e.g. languages with a number of grammatical cases, articles or capitalisation of nouns. Whisper, which performed best in our study, reports an average multilingual \ac{WER} that is three times higher compared to the English \ac{WER} \cite{Whisper2022}. This discrepancy in performance is further highlighted considering the spread reported by Whisper on the Common Voice 15 dataset, which ranges from 4.3\% to 55.7\% \ac{WER}. Even though \ac{ASR} might meet the accuracy requirements for caption users for English, it might fail to meet them for other languages and outside of narrowly defined scenarios, which is a real problem for the accessibility of events.

Our English video recordings had a similar \ac{WER} as the samples taken from the LibriSpeech corpus. Most of the tested \ac{ASR} services seem to have a robust general performance, and do not only achieve particularly good results on a common dataset. An evaluation on LibriSpeech can reflect the general accuracy of an \ac{ASR}-model, if it is not explicitly optimised for this dataset. We still see the danger that extremely low error rates reported on common datasets cause us to misjudge general \ac{ASR} performance. This might also be a challenge for the regulation of \ac{ASR} as requested by the \ac{NAD} \cite{NADPetition2019}: Setting a threshold on a publicly available dataset could cause a model to be overly adapted to that dataset, in order to meet legal requirements.

Whisper as an open-source model outperforms commercial services in many cases and reports quite low error rates in general. These findings are also reported by other studies \cite{Whisper2022, AssemblyAI2023, Deepgram2022, SpeechmaticsUrsa2023}. This is a strong indicator that \ac{E2E} models can further increase the accuracy of \ac{ASR}. However, accuracy is not the only relevant factor for \ac{ASR}, and \ac{E2E}-models have shown weaknesses in streaming capabilities, latency and computational efficiency \cite{Li2022}. They also tend to be inaccurate in predicting word-level timestamps \cite{Bain2023}. The strength of an open source model like Whisper is its ability to support further scientific research, and to be adapted to specific use-cases such as Higher Education.

Adding vocabularies did not significantly improve the accuracy of the different \ac{ASR} services. However, some technical terms, abbreviations and names did only appear in the transcript when a vocabulary was provided. Even though adding the vocabulary did not significantly decrease the overall \ac{WER}, these words might be fundamental for understanding the content and might consequently make the transcript more accessible. Presenters or speakers should consider the use of vocabularies if their text contains abbreviations or specialist vocabulary, even though there is no guarantee that any of these words will be correctly identified and transcribed by an \ac{ASR}-model. Ideally, such vocabularies can be created with little effort, for example by automatically extracting rare words from presentation slides.

We also found no indication that the processing duration affects the resulting accuracy. Longer processing time does not necessarily result in a higher accuracy and most likely results from the available computing resources. One method to increase the processing speed is to split the recording into short clips, transcribe these segments concurrently, and then merge the transcribed texts. A related study reported a speed increase of up to twelve times without sacrificing transcription accuracy \cite{Bain2023}. It is very likely that Deepgram uses such an approach, given its very fast performance in our evaluation.

Streaming \ac{API}s showed a significantly higher error rate compared to batch transcription. To achieve near real-time transcription, a low \ac{RTF} is required to reduce the latency between audio and transcription results. Most probably, vendors have a separate \ac{ASR}-system for streaming, which sacrifices some accuracy to increase speed. Hybrid \ac{ASR}-models could use a different processing pipeline and \ac{E2E}-models could use a smaller model size, as larger models generally take longer to process. However, as \ac{ASR} is particularly useful for real-time events, accuracy should also be evaluated with the corresponding models, as long as there is a significant difference to non-streaming models.

The confidence outputs of \ac{ASR} services were not transferable to the accuracy of a transcript. The confidence values merely seem to represent the probabilities within an \ac{ASR}-model, and not the actual correctness. Visually highlighting words regarding their confidence, as explored by \citet{Seita2018} might cause more confusion for the readers than it actually helps. It is also questionable if the confidence outputs can be used as part of an error prediction of \ac{ASR}, as suggested by \citet{Kafle2016}.

Alternative error metrics showed similar results compared to the \ac{WER} and seem to be interchangeable. The \ac{CER} may be useful for evaluating languages that do not separate words with spaces, such as Chinese \cite{Whisper2022}. It is common in speech recognition research to apply text normalisation to make the \ac{WER} more reliable to formatting differences. Without these modifications the average error rate in our study was around twice as high. The extended normalisation that replaces common contractions and unifying numbers showed a significant reduction of the \ac{WER}, compared to only lowercase transformation and the removal of punctuation. These normalisations should therefore be used for metrics, that are based on string edit-distances, to better reflect the amount of content errors. However, institutions like the \ac{FCC} demand correct capitalisation and punctuation of captions to aid comprehension \cite{FCC2014}. As the \ac{WER} is less meaningful without text normalisation, we need additional metrics that reflect the accuracy of punctuation, capitalisation and other aspects such as formatting.

Complementary to large scale automated evaluations and the search for additional metrics, it is crucial to evaluate \ac{ASR} qualitatively, especially with \ac{DHH} individuals. User studies can identify additional issues with \ac{ASR}, such as the readability of \ac{ASR}-generated text \cite{Butler2019} or the impact of latency in interactive conversations \cite{McDonnell2021}. Focus groups can also provide ideas on how to improve the use of \ac{ASR} \cite{Seita2022}, or help assess the acceptance of new features like automatic content summarisation \cite{Alonzo2020}. Acceptance of \ac{ASR} as an accessible accommodation is an important factor, as it can be perceived as a second-rate service by users \cite{NDC2020}. Human involvement, e.g. by correcting or monitoring \ac{ASR} \cite{McDonnell2021, Seita2022}, can potentially increase the acceptability of \ac{ASR} independent of its actual accuracy.

\section{Conclusion}

We provided a comprehensive overview on state-of-the-art quality of many common \ac{ASR} services, that puts extremely low error rates reported by scientific research and general claims by vendors into perspective. We could confirm very high accuracy rates in some cases, but found a wide range in quality across vendors, speakers and languages. The range of error rates in our dataset shows that average accuracy rates can be misleading. This rather weak reliability must be considered, if \ac{ASR} is used as an accessibility tool without human supervision.

We also found a significantly higher error rate for streaming \ac{ASR}. As automatic transcription is particularly useful for real-time events, this performance gap must be closed. Technical enhancements like an additional vocabulary showed only little improvement. Also the usage of \ac{ASR} metadata like the word level confidence to indicate errors is doubtful.

With the increasing availability of \ac{ASR} in commercial tools, it is already part of our social and work life. But only because \ac{ASR} offers a solution for the complex task of transcription, it is not necessarily accessible and enables \ac{DHH} individuals to participate in various events. As demanded by the \ac{DHH} community, we need a declaratory ruling on the use of \ac{ASR} for transcription, which largely depends on a binding metric and quality threshold. Despite criticism that the \ac{WER} does not necessarily reflect the usefulness of a text \cite{Wang2003, Mishra2011, Favre2013}, it still holds potential as a complementary accessibility metric. Primary because it is objective, unbiased, and commonly used as an evaluation metric in \ac{ASR} research. The \ac{WER} is particularly useful for verbatim transcription, and extensive text normalisation can reduce errors caused by non-semantic differences. \ac{ASR} continues to improve and error rates decrease, making metrics based on a string edit distance more robust. However, it cannot be used exclusively, as text normalisation removes important aspects of punctuation or capitalisation, and other factors such as speaker diarisation or caption formatting are not covered.

In many situations, speakers may not be able to choose a particular or even pre-trained \ac{ASR} system, as it will be determined by the meeting or presentation software. However, speakers can be more inclusive when presenting to diverse audiences by speaking more clearly and slowly to give readers time to adjust to transcription errors. A prior study indicates that speakers adjust their behaviour in meetings with \ac{DHH} peers when \ac{ASR} is used \cite{Seita2018}.

\section{Limitations}

Our objective was to provide a realistic picture of the accuracy of \ac{ASR}, but the generalisability of our results is limited. The setting we chose, Higher Education, provided rather ideal conditions and avoided particular weaknesses of \ac{ASR}. However, \ac{DHH} individuals are confronted with less ideal scenarios in real life. Results may differ in other environments or for more conversational speech. Furthermore, our dataset reflects biases inherent in the material available online and the over-representation of certain groups in Higher Education. The findings are also restricted to the date of the evaluation, as \ac{AI} is a rapidly changing area of research. Due to the large amount of data, the analysis relies on the \ac{WER} as the most common automated metric, which is a limited representation of \ac{ASR} quality.

\section{Future Work}

Although this study provides only a snapshot of \ac{ASR} quality, and performance will certainly improve in the future, many of the problems that caption users currently face are likely to remain. Our chosen scenario avoided many of the challenges of \ac{ASR}, such as speaker diarisation, overlapping speech and mixed language. Furthermore, \ac{ASR} performs notably better for English and German compared to many other languages. Evaluations of \ac{ASR} in more challenging environments and for underresourced languages may find considerably higher error rates and less useful transcription results. Increasing the amount of training data can effectively improve the overall performance of an \ac{ASR}-model and its robustness to specific biases, such as accented speech, or the speech of the elderly or children. For \ac{ASR} to be truly inclusive, it is also important to address the challenges of transcribing speech from people with speech impairments such as fluency disorders or dysarthria.

Due to the variety of applications of \ac{ASR}, a single metric appears insufficient to comprehensively represent the quality of an \ac{ASR}-model. Currently, the \ac{WER} serves as the standard metric for reporting accuracy, but because text normalisation is used to reduce non-semantic differences, information about the correctness of punctuation and capitalisation is lost. A future metric could provide a more detailed analysis that distinguishes between error types and reports accuracy on several aspects. In addition to punctuation and capitalisation, the type of word errors, e.g. technical terms, numbers, abbreviations, homonyms, could be classified according to their impact on text comprehension. Ideally, empirical evaluations can establish a correlation between such an automated metric and subjective ratings.

\begin{acks}
We thank Andreas Burkard for his efforts in making this paper accessible and Kathy-Ann Heitmeier for her contributions to English language refinement. This work was conducted as part of the SHUFFLE Project (“Hochschulinitiative Barrierefreiheit für alle”) and funded by “Stiftung Innovation in der Hochschullehre”.
\end{acks}

\bibliographystyle{ACM-Reference-Format}
\bibliography{bibliography}


\end{document}